\documentclass{article}
\usepackage{graphicx} %
\usepackage{parskip} %
\usepackage{cprotect} %
\usepackage{biblatex} %
\usepackage{hyperref} %
\usepackage{caption} %
\usepackage{subcaption} %
\usepackage{authblk} %

\hypersetup{
    colorlinks=true,
    citecolor=cyan
}

\title{Block-level Text Spotting with LLMs}
\author{Ganesh Bannur, Bharadwaj Amrutur}
\affil{Indian Institute of Science}
\date{}

\begin{document}

\maketitle
\newcommand{\technique}{\textsc{BTS-LLM}}
\newcommand{\subscript}[2]{\texorpdfstring{#1\textsubscript{#2}}{$#1_#2$}}

\begin{abstract}
    Text spotting has seen tremendous progress in recent years yielding performant techniques which can extract text at the character, word or line level. However, extracting blocks of text from images (block-level text spotting) is relatively unexplored. Blocks contain more context than individual lines, words or characters and so block-level text spotting would enhance downstream applications, such as translation, which benefit from added context. We propose a novel method, \technique{} (\textbf{B}lock-level \textbf{T}ext \textbf{S}potting with \textbf{LLM}s), to identify text at the block level. \technique{} has three parts: 1) detecting and recognizing text at the line level, 2) grouping lines into blocks and 3) finding the best order of lines within a block using a large language model (LLM). We aim to exploit the strong semantic knowledge in LLMs for accurate block-level text spotting. Consequently if the text spotted is semantically meaningful but has been corrupted during text recognition, the LLM is also able to rectify mistakes in the text and produce a reconstruction of it. %
\end{abstract}

\section{Introduction}
    \subsection{Text Spotting at Different Levels}
\begin{figure}
            \centering
            \makebox[\textwidth][c]{\includegraphics[width=1.25\textwidth]{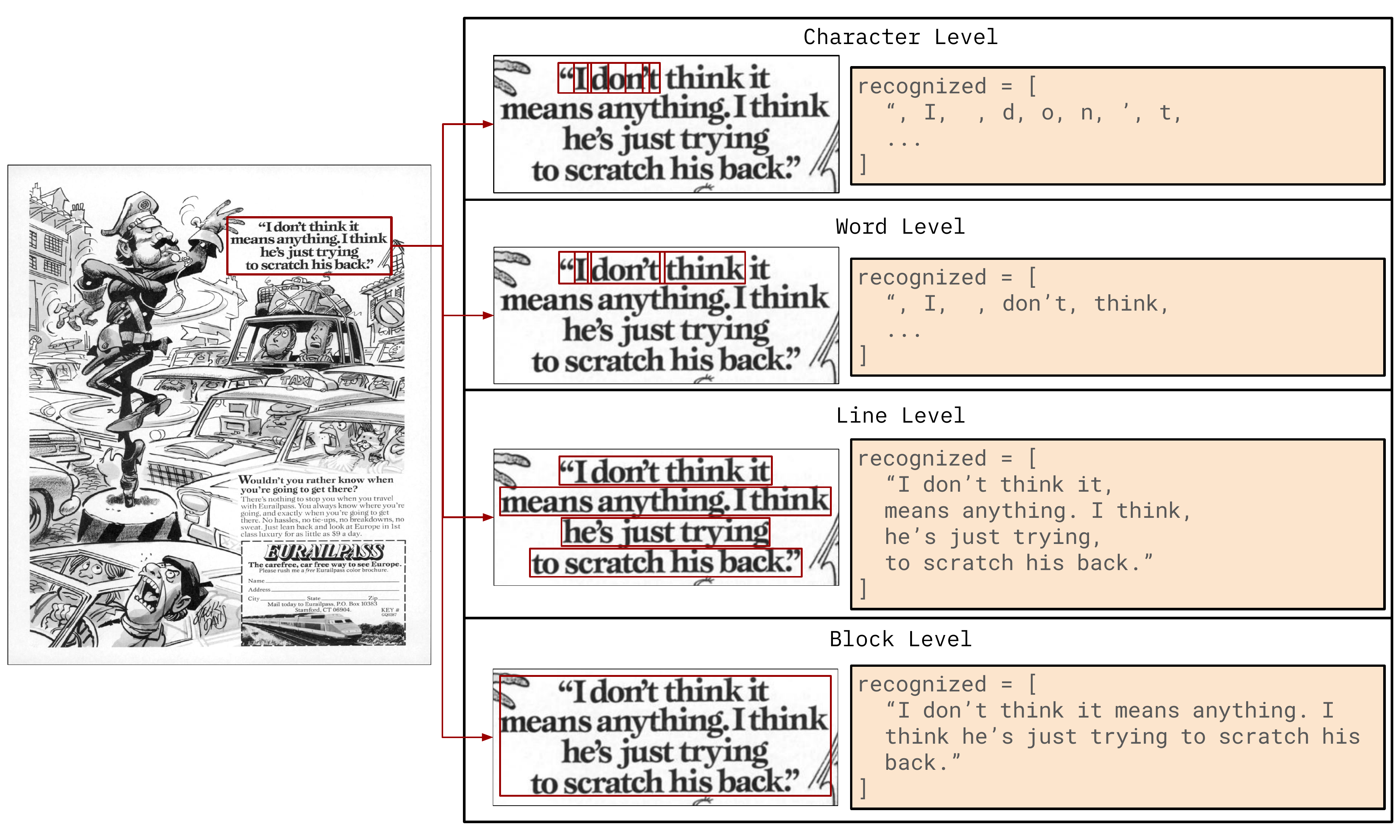}}
            \caption{The four levels of text spotting}
            \label{fig:spotting_levels}
        \end{figure}

        Text spotting is the term used to describe techniques which extract text present in natural images (as opposed to document images). There are two types of techniques for text spotting: 1) end-to-end and 2) pipelined. End-to-end techniques consist of a single text spotting model or jointly trained pipeline which can both detect and recognize text. Pipelined techniques break the task down into text detection and text recognition which are solved by separate models. Currently both end-to-end and pipelined state-of-the-art techniques work at the character, word or line level while block-level text spotting has remained relatively unexplored. We propose a technique which can extract text from natural images at the block level. Figure \ref{fig:spotting_levels} shows the difference between the four levels of text spotting.
        
        Recently end-to-end approaches have gained popularity since using a single model for detection and recognition allows for information sharing between the tasks, which improves results. However we choose a pipelined approach for two reasons: 1) Using a pipelined approach allows us to use the Unified Detector from \cite{long2022endtoend} which enables the first step in creating blocks from lines (described in Section \ref{subsection:ltswg}). 2) There are instances showing that pipelined approaches may still perform better than end-to-end approaches. An example is the ICDAR Competition on Hierarchical Text Detection and Recognition \cite{long2023icdar} in which all the top submissions were pipelined. %

    \subsection{The Challenge of Block-level Text Spotting} \label{subsection:why_llm}
\begin{figure}
            \centering
            \makebox[\textwidth][c]{\includegraphics[width=1.25\textwidth]{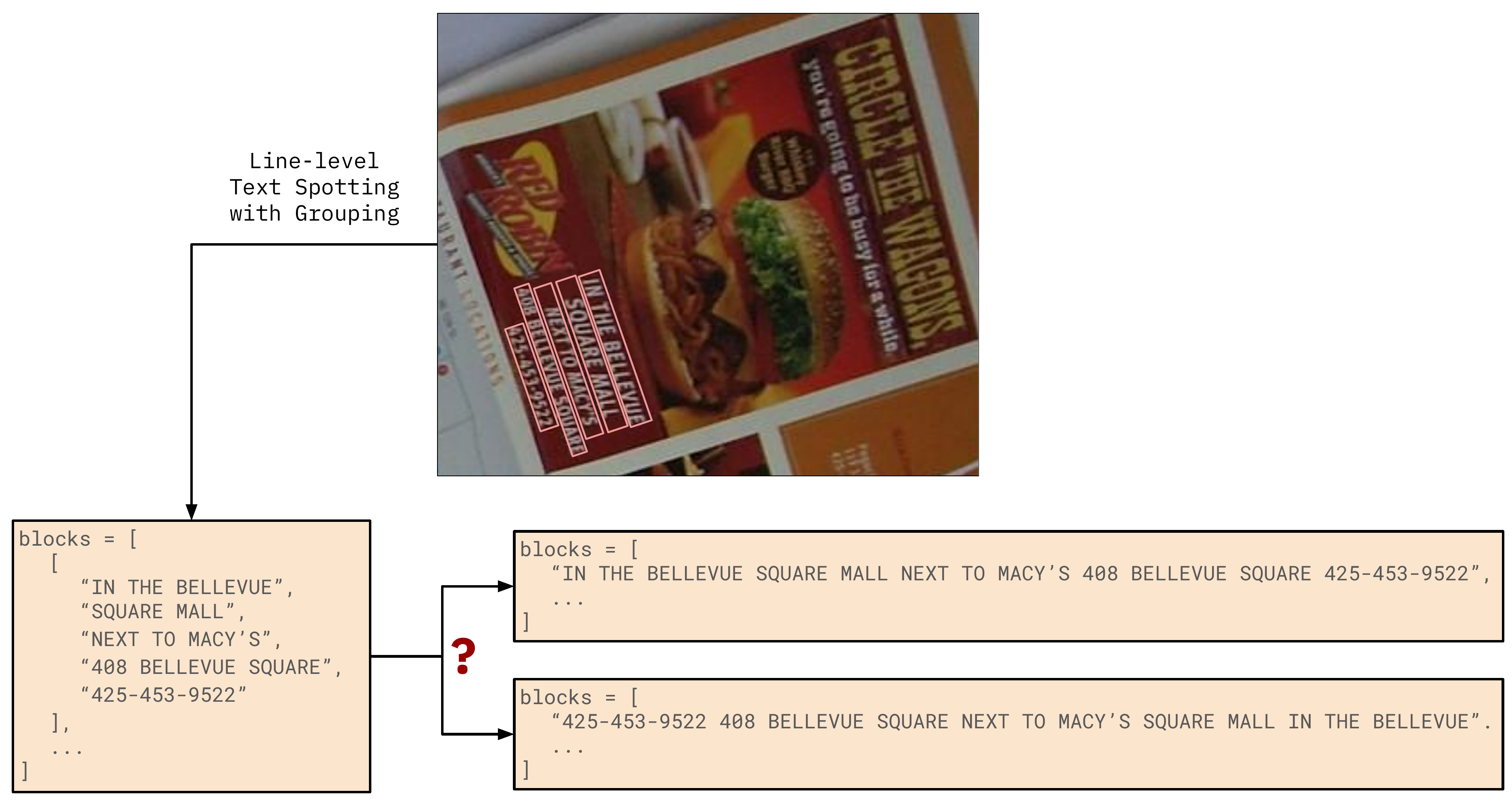}}
            \cprotect\caption{A set of lines identified as being part of the same block are highlighted in the image. The first part of the pipeline will identify the bounding box and the corresponding text of each line. Based on their bounding boxes there are two ways to order these lines, which are shown as the two possibilities for the array \verb|blocks|. The correct order is the one on top however if we were to simply read left-to-right in the image we would pick the bottom order. To know which one to pick it becomes important to read the text and decide which order makes more sense.}
            \label{fig:arrangement_possibilities}
        \end{figure}

        Using its pipelined approach, \technique{} is able to extract the following: 1) The bounding box of every line, 2) The corresponding text of every line and 3) The grouping of lines into blocks. The next step is to find the text of each block. Since the lines are already grouped into blocks, the main task in creating the text for each block is ordering the lines. Figure \ref{fig:arrangement_possibilities} demonstrates the challenge of ordering the lines. %
        
        As Figure \ref{fig:arrangement_possibilities} shows, when ordering lines it is important to understand the meaning of the text in addition to understanding the arrangement of bounding boxes. This presents two challenges: 1) The text identified has an extremely broad scope since it can be part of any scene image and 2) Any method to identify the correct order would need to understand both the arrangement of bounding boxes and the meaning of text, which presents a complex spatio-linguistic task. LLMs are able to understand an extremely wide scope of text. They have also been shown to be able to perform generic reasoning in textual and visual modalities \cite{bubeck2023sparks} and have been shown to perform well on 2D spatial reasoning \cite{sharma2023exploring}. Therefore we propose to use an LLM to perform the ordering as it can meet the challenges.

    \subsection{Benefits of Incorporating an LLM}
        The last step in text spotting is scene text recognition (STR), which "reads" the text. STR is not purely a vision task but rather a vision-language task. Semantic knowledge can be used to augment the visual features and improve STR results. Recent work in STR has focused on improving the language modeling capabilities in STR models and has consequently improved text recognition performance \cite{bautista2022scene}. While language models (LMs) have improved STR results, they are constrained by their size and training data. The LMs used in STR are small and are only trained on the text labels seen in STR datasets, thus limiting their semantic knowledge. In recent years language modelling has been disrupted by the rise of LLMs. We posit that STR could benefit from the significantly larger model size and more general training data of LLMs. While \technique{} uses an LLM to order lines within a block, the vast semantic and world knowledge embedded in LLMs could also improve its text spotting results.

    \subsection*{Contributions}
    Our contributions are as follows:
    \begin{itemize}
        \item We provide a pipelined approach for block-level text spotting.
        \item We incorporate an LLM into text spotting. We posit that this improves text spotting quality by allowing for reconstructing semantically meaningful text that has been corrupted during text recognition.
    \end{itemize}

\section{Related Work}
    \subsection{Block-level Detection, Recognition and Spotting}
        \subsubsection{Techniques For Scene Text}
            The techniques closest to our work fall under this category. \cite{wei2022textblock, long2024hierarchical} propose block-level text spotting techniques. \cite{wei2022textblock} has a block-level detector followed by a block-level recognition model which can directly recognize the text in a block. In \cite{wei2022textblock} the block-level training dataset is heuristically created from line-level datasets by merging lines into blocks based on Intersection-over-Union (IoU) values. \cite{long2022endtoend} propose a block-level text detection technique in which lines are detected and grouped into blocks. \cite{long2024hierarchical} builds upon \cite{long2022endtoend} and proposes a block-level text spotting technique which performs line-level spotting and then groups lines into blocks. However the order of lines inside a block is not determined in both \cite{long2022endtoend} and \cite{long2024hierarchical}. \cite{xue2022contextual} proposes a block-level text detection technique. They detect individual text units (usually words), group them into blocks and predict the reading order within each block.
        \subsubsection{Block-level Text Detection For Documents}
            Over the years, there has been a lot of interest in identifying and extracting blocks of text within structured documents. Early works usually used statistical techniques to create text blocks by clustering word or line bounding boxes \cite{chen1996extraction}. Recent approaches apply deep learning architectures (usually graph convolutional networks) on bounding boxes to create blocks \cite{liu2022unified, wang2022post}. Most block-level detection techniques for documents focus on only the bounding boxes and ignore the actual text in words or lines. The Unified Detector from \cite{long2022endtoend}, which we use to detect blocks, is inspired by the field of block-level text detection for documents.
        \subsubsection{Block-level Text Recognition For Handwritten Text}
            There has been a substantial amount of work done in recognizing blocks of handwritten text \cite{bluche2016joint, bluche2017scan, chung2019computationally, yousef2020origaminet, coquenet2021span}. Most of the work done for handwritten text is specifically in block-level recognition. The focus has been on recognizing text in a cropped image of a single paragraph or in an image of a full page of handwritten text (usually without anything else, such as images, on the page). The text to be recognized is generally on a plain background and with blocks consisting of a column of lines.
    
    \subsection{Reading Order Detection}
        \technique{} uses an LLM to order lines within a block, that is, it finds the reading order of the lines. While there are no reading order detection techniques for scene text, the document domain has seen an evolution of reading order detection techniques over time. Reading order detection started with geometric methods which applied graph sorting \cite{aiello2003bidimensional} or optimization \cite{meunier2005optimized} algorithms on the bounding boxes of text blocks. Techniques then evolved to use learning algorithms to adapt to new classes of documents by learning the domain specific layout knowledge for each new class of documents \cite{ceci2007data, malerba2008machine, quiros2022reading}. Others have also tried to impose document-agnostic generic constraints on text blocks to narrow down and find a reading order \cite{ferilli2014abstract}. Recent techniques have used deep learning architectures which incorporate both layout and textual information to find the reading order \cite{wang2021layoutreader}. However block-level text spotting involves scene text which is out of domain for these document techniques. Scene text is also a more complex domain than documents for reading order detection due to the free structure of text in scenes. %

\section{Method}
\begin{figure}
        \centering
        \makebox[\textwidth][c]{\includegraphics[width=1.25\textwidth]{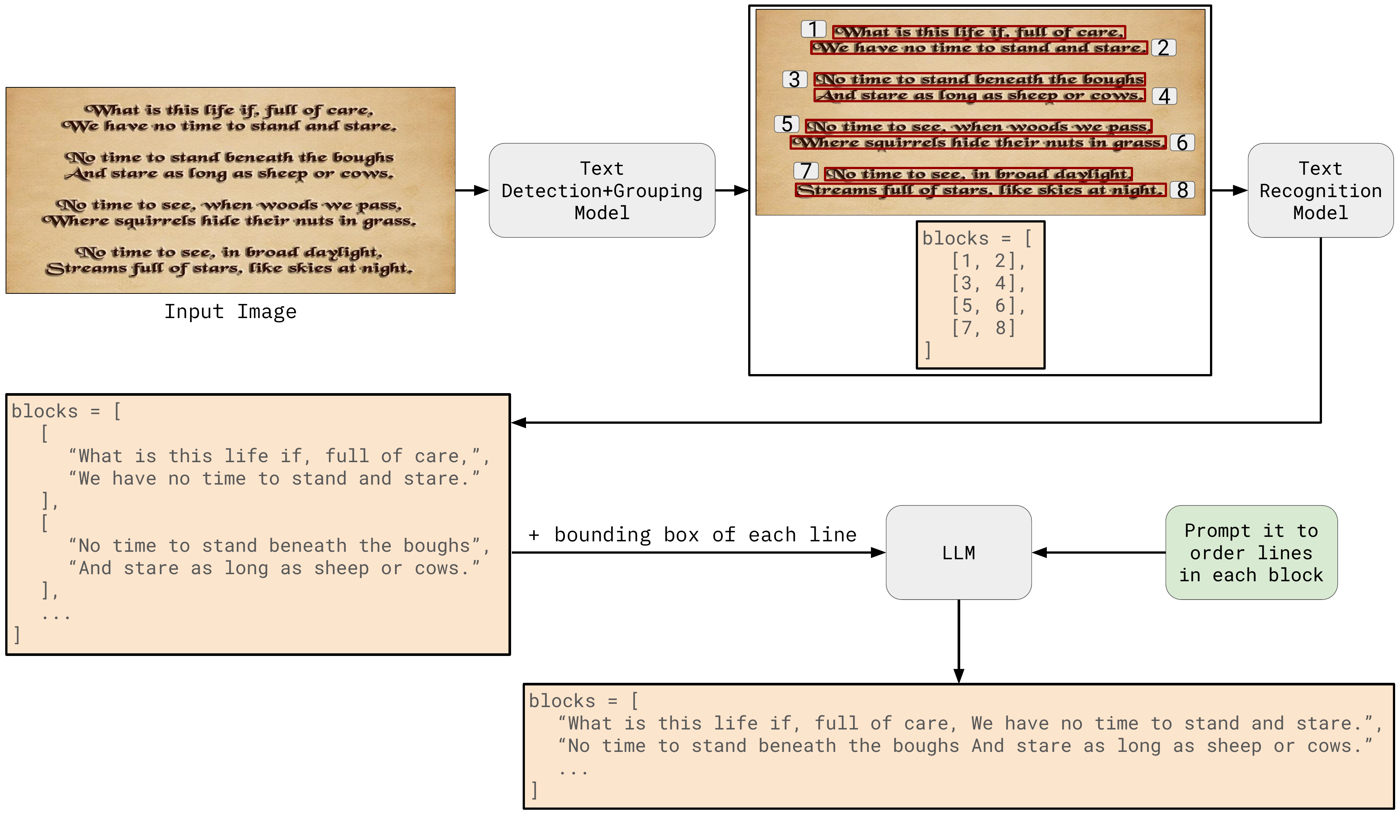}}
        \cprotect\caption{The pipeline of \technique{} is as follows. An input image is given to the Text Detection+Grouping Model. This model performs line-level text detection and finds the bounding boxes of all the lines in the image. It then groups lines in close proximity into blocks. In the figure, the lines detected are numbered from 1 to 8 and the grouping is shown in the array \verb|blocks|. The detected regions are given to the Text Recognition Model which recognizes the line of text in each region. Finally, for every block, the texts recognized and the bounding boxes of the lines are given to the LLM which outputs the text for the block.}
        \label{fig:pipeline}
    \end{figure}

    Figure \ref{fig:pipeline} shows the overall pipeline of \technique{}. The two mains parts of the pipeline are: 1) Line-level text spotting with grouping (covered in Section \ref{subsection:ltswg}) and 2) Creating blocks using an LLM (covered in Section \ref{subsection:cbual}). The pipeline utilizes pre-existing state-of-the-art techniques as its components.
    
    \subsection{Line-level Text Spotting with Grouping} \label{subsection:ltswg}
        As shown in Figure \ref{fig:pipeline}, the first model in the pipeline is the Text Detection+Grouping Model which takes an input image and does two things: 1) It performs line-level text detection and identifies the bounding boxes around each line in the image. 2) It groups lines which are in close proximity as being part of the same block. The image shows 8 lines numbered 1 to 8 whose grouping is shown in the array \verb|blocks|. Lines 1 and 2 are in close proximity and so are grouped as being part of the same block and similarly for the other lines. The model used for text detection+grouping is the Unified Detector from \cite{long2022endtoend} (note that blocks are called paragraphs in \cite{long2022endtoend}).

        Next the image of each line is preprocessed and given to the Text Recognition Model which recognizes the text in the cropped image. The PARSeq model from \cite{bautista2022scene} is used for text recognition. Preprocessing includes the following steps performed for each bounding box:
        \begin{enumerate}
        \item The image is rotated to align the rectangular bounding box with the axes. If the box is rotated by $\le$45\textdegree then it is rotated back to align with the x-axis. Otherwise it is rotated to align with the y-axis. This may result in the images of some lines being rotated by 90\textdegree, however as Tables \ref{table:PARSeq_90}, \ref{table:PARSeq_180} and \ref{table:PARSeq_270} show, PARSeq is robust to rotation.
        \item The rotated image is cropped to the bounding box.
        \item The cropped image is split horizontally into parts that maintain the input aspect ratio of the text recognition model (for PARSeq the aspect ratio is height$\times$width = 32$\times$128). This is done since any input image is first converted to this aspect ratio before being fed into the model. If the actual aspect ratio of the image varies a lot from the input aspect ratio then the text in the image appears squished and the accuracy of the model decreases. This is prevented by having each part maintain the input aspect ratio. Each part is then fed into the model separately and the strings recognized from each part are concatenated to get the text for the line.
        \end{enumerate}
        Since the model released in \cite{bautista2022scene} cannot recognize spaces, we retrained it to recognize spaces. Tables containing results from the retrained model are given in Appendix \ref{appendix:rorpm}.
        
        Finally, this part of the pipeline outputs the following: 1) the bounding box of every line, 2) the corresponding text of every line and 3) the grouping of lines into blocks.

    \subsection{Creating Blocks using an LLM} \label{subsection:cbual}        
\begin{figure}
            \centering
            \makebox[\textwidth][c]{
            \begin{tabular}{c}
            \begin{subfigure}[c]{1.25\textwidth}
                \centering
                \includegraphics[width=\textwidth]{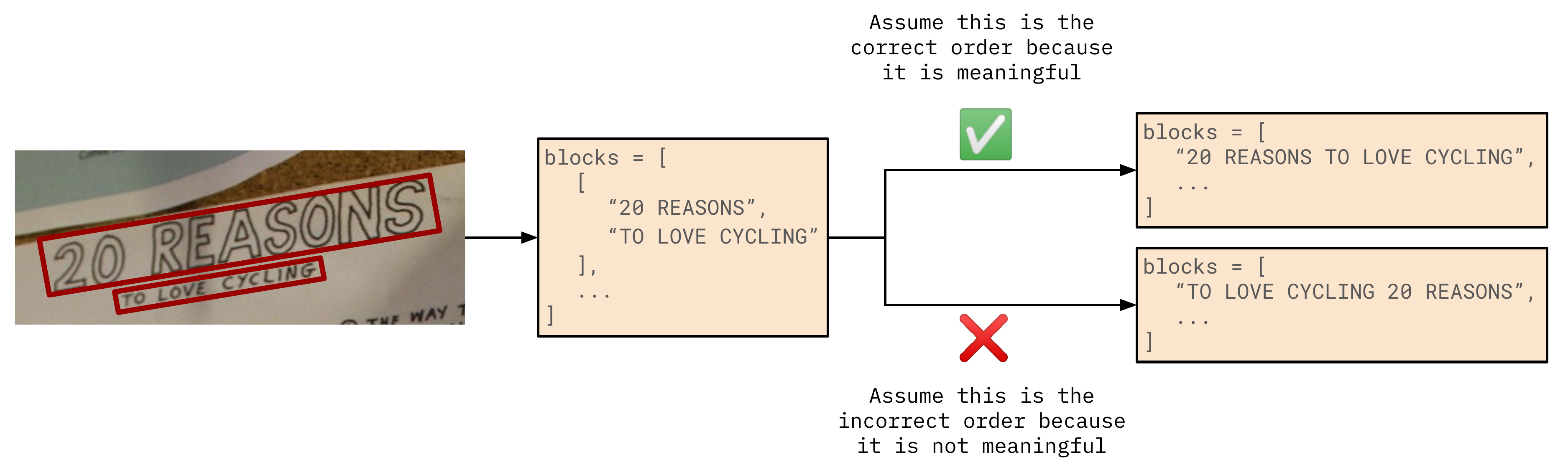}
                \caption{Category 1}
                \label{fig:assumption:category_1}
            \end{subfigure}\\ \\
            \begin{subfigure}[c]{1.25\textwidth}
                \centering
                \includegraphics[width=\textwidth]{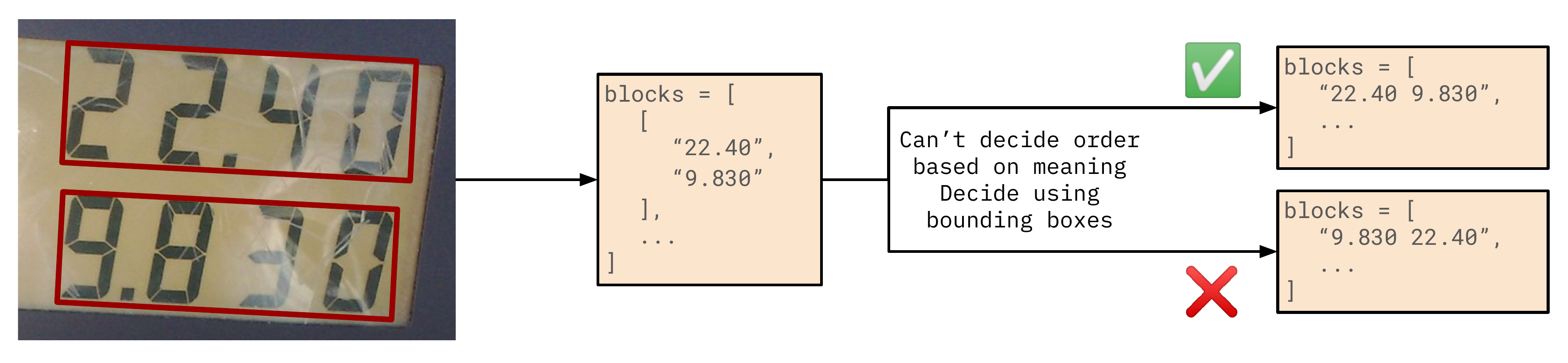}
                \caption{Category 2}
                \label{fig:assumption:category_2}
            \end{subfigure}
            \end{tabular}}
            \caption{We assume that blocks fall into one of two categories}
            \label{fig:assumption}
        \end{figure}

        At this point in the pipeline, the lines have already been recognized and grouped into blocks. The main task in obtaining the text for each block is ordering the lines. As detailed in Section \ref{subsection:why_llm}, we use an LLM for this task. Using an LLM for a task requires the formulation of a prompt. Before discussing the prompt it is necessary to discuss an assumption . We assume that all blocks encountered fall into two categories, which are illustrated in Figure \ref{fig:assumption}.

        \begin{enumerate}
            \item \textit{Category 1: Blocks for which the meaningful order is the correct order}\\
            If there is a meaningful order then we do not look at bounding boxes and assume that the meaningful order is the correct order. In other words, we prioritize semantics over spatial arrangement. There may be rare cases in which this is not true and the meaningless order is the correct order. For example, if in Figure \ref{fig:assumption:category_1} the line \textit{"TO LOVE CYCLING"} was above the line \textit{"20 REASONS"} the correct order would be \textit{"TO LOVE CYCLING 20 REASONS"} but \technique{} would give \textit{"20 REASONS TO LOVE CYCLING"} since it is the meaningful order.
            \item \textit{Category 2: Blocks for which none of the orders are meaningful}\\
            If none of the orders are meaningful then the lines are ordered based on bounding boxes. In other words, we use spatial arrangement when semantics fail.
        \end{enumerate}
        In the corner case when more than one order is meaningful, the LLM will decide which one to choose.
        
\begin{figure}
            \centering
            \makebox[\textwidth][c]{\includegraphics[width=1.25\textwidth]{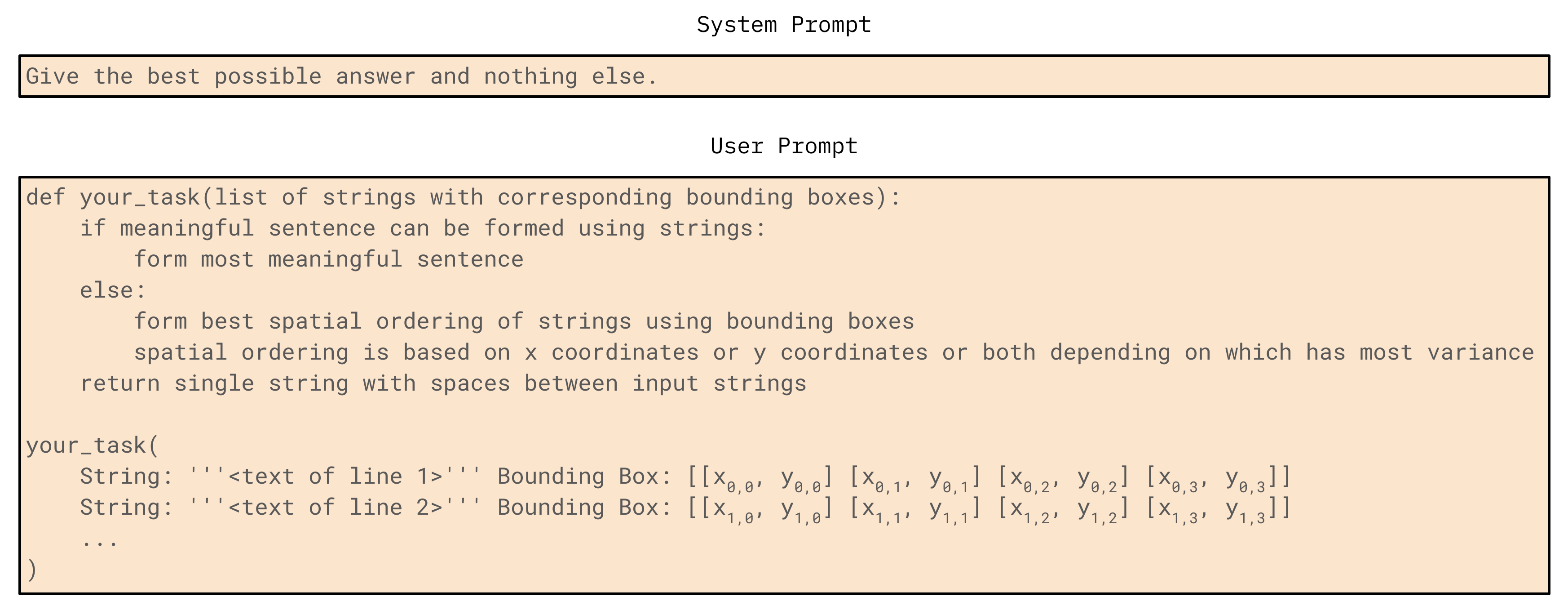}}
            \cprotect\caption{The prompt provided to the LLM. \technique{} uses GPT-3.5 Turbo which requires a system prompt. For other LLMs which do not have a system prompt, it can be appended before the user prompt. The system prompt given was empirically found to be the best for two reasons: 1) It suppresses extraneous output. 2) It forces GPT-3.5 Turbo to always give an answer (the best possible one, given the input). Finally, leaving the system prompt blank was seen to severely decrease the quality of outputs. GPT-3.5 Turbo is accessed via the API\footnotemark with \verb|temperature=0|.}
            \label{fig:prompt}
        \end{figure}

        Having made this assumption the task can be described as follows: "For every block, if there is a meaningful order of lines then output that order. Otherwise use the bounding boxes to order the lines". The prompt needs to encode this if-else construct. The best way to do this and also specify the task very precisely was to formulate the prompt as code and make the LLM predict the output of this code. However we cannot write code for subtasks such as finding a meaningful order of lines. Instead we interleave code with natural language instructions, taking advantage of 1) the LLM's ability to understand code and follow the precise constructs, and 2) its ability to perform tasks specified in natural language. The prompt is formulated in two parts: 1) Function definition for the function \verb|your_task|, which defines the task by interleaving code and natural language. 2) Function call for \verb|your_task| with the inputs passed as function arguments, which prompts the LLM to produce the output of \verb|your_task| on those inputs. Python is used for code since it has the least amount of extraneous characters (for example semicolons at the end of each line in C/C++/Java) and is one of the more readable general purpose languages. The prompt given to the LLM is given in Figure \ref{fig:prompt}. The LLM is executed with this prompt for every block to get the corresponding text. An additional detail of the prompt is the order in which lines are given as function arguments. To keep the order consistent, the lines are first ordered either top-to-bottom-left-to-right or left-to-right-top-to-bottom using a heuristic-based geometric ordering algorithm which orders using bounding boxes. The lines are then passed as arguments in this order.

        \footnotetext{\url{https://platform.openai.com/docs/models/gpt-3-5-turbo}}

        Our approach to prompt design is similar in spirit to \cite{li2023chain}. In \cite{li2023chain} ambiguous tasks are specified as function calls whose function name describes the task (\cite{li2023chain} gives the example of \verb|detect_sarcasm(string)|). In our approach, however, ambiguous tasks (such as finding a meaningful order) are directly specified in natural language. Empirically this could have two benefits: 1) It is clearer for the LLM which parts are code (which must be executed precisely) and which parts are ambiguous tasks (which must be answered using general knowledge and reasoning). 2) Describing ambiguous tasks with natural language allows for a better description of the task rather than inferring the task from a function name. Note that in \cite{li2023chain} it is actually the LLM which produces the text which interleaves code and natural language following which only the ambiguous parts are executed by an LLM.
        
        \subsubsection*{Additional Aspects of using an LLM} \label{subsubsection:additional_aspects}
        
            The first consideration is the values of bounding box coordinates provided in the prompt. To decrease the numerical values of the coordinates they are translated such that the minimum x and y coordinates among all bounding boxes in a block are 0. Having smaller numbers makes it easier for the LLM to perform numerical reasoning as there is a higher chance of small numbers occurring in the training data of the LLM, making them "in-distribution" and improving the LLM's performance on them. For the same reason, the coordinates are kept as integers and are not normalized to [0, 1] since small (0 to $\approx$1000 for bounding box coordinates) positive integers have a higher chance of occurring in the training data than decimals. Keeping numbers in-distribution is important since the transformer architecture of LLMs suffers from length generalization in numerical and reasoning tasks \cite{awasthi2023improving}.
            
            The second consideration is the context length of the LLM. If the prompt exceeds the context length then \technique{} does not use the LLM to order lines but rather defaults to using the heuristic-based geometric ordering mentioned previously. The lines are ordered either top-to-bottom-left-to-right or left-to-right-top-to-bottom based on only their bounding boxes.
    
            Finally, LLMs, being probabilistic next token predictors, are prone to mistakes due to the sampling method used. Sampling mistakes can affect the length of the generated output. To curtail this problem, if the length of the LLM's output is less than half or more than double the expected length then \technique{} defaults to using the heuristic-based geometric ordering mentioned previously. The expected output length for a block is the length of the string formed by concatenating the line texts with spaces inserted between them. \technique{} uses GPT-3.5 Turbo with \verb|temperature=0| to 1) make outputs more deterministic and 2) focus the LLM on the input data rather than generating new text. This makes it even more important to impose bounds on output length since the problem of text degeneration and repetitive outputs are exacerbated at low temperature settings \cite{holtzman2020curious} (this is true for both temperature and nucleus sampling provided by the GPT-3.5 Turbo API; see Figure 9 in \cite{holtzman2020curious}).

\section{Evaluation}
    \subsection{Dataset}
        For evaluation, we create a block-level text spotting dataset using the validation set from the Hiertext dataset \cite{long2022endtoend} (the test set has not been released). The Hiertext dataset is a line-level dataset however it provides grouping of lines into blocks. We manually ordered the lines in each block to create the block-level dataset. Blocks for which the reading order was not apparent to a human were not considered for evaluation.
        
    \subsection{Metrics}
        The quality of the text recognized for each block is measured by finding the similarity between the predicted string and ground truth string for each block. To gain a comprehensive understanding we use a wide variety of similarity metrics. The metrics used are:
        \begin{itemize}
            \item \textit{Learned Metrics}: Sentence BERT (cosine similarity between Sentence BERT embedding of prediction and ground truth), BERTScore, BLEURT\\
            These metrics capture semantic similarity between predictions and ground truth. We use these metrics to judge the semantic likeness between predicted and ground truth strings.
            \item \textit{String Similarity Metrics}: Normalized Levenshtein distance, Jaro-Winkler similarity, Ratcliff-Obershelp similarity\\
            These traditional string similarity metrics capture the character-level likeness between predicted and ground truth strings.
        \end{itemize}
            
    \subsection{Protocol}
        We are interested in comparing the predicted text for a block with the ground truth and are not interested in measuring the text detection quality (see \cite{long2022endtoend} for an evaluation of Unified Detector, \technique{}'s text detector). Hence each predicted bounding box is matched to the ground truth bounding box with which it has the maximum IoU. Since one ground truth bounding box can have multiple predicted bounding boxes matched with it, we need to find the part of the ground truth string corresponding to each predicted string in order to compare the two. For each predicted string, we find its best fuzzy substring match in the ground truth string (using the algorithm in Appendix \ref{appendix:substring}). The best fuzzy substring match of a string \verb|query| in a string \verb|corpus| is defined as the substring of \verb|corpus| which has the lowest Levenshtein distance from \verb|query|. Metrics are then calculated using the predicted string and its best fuzzy substring match.

    \subsection{Results}        
\begin{figure}
            \centering
            \makebox[\textwidth][c]{\includegraphics[width=1.25\textwidth]{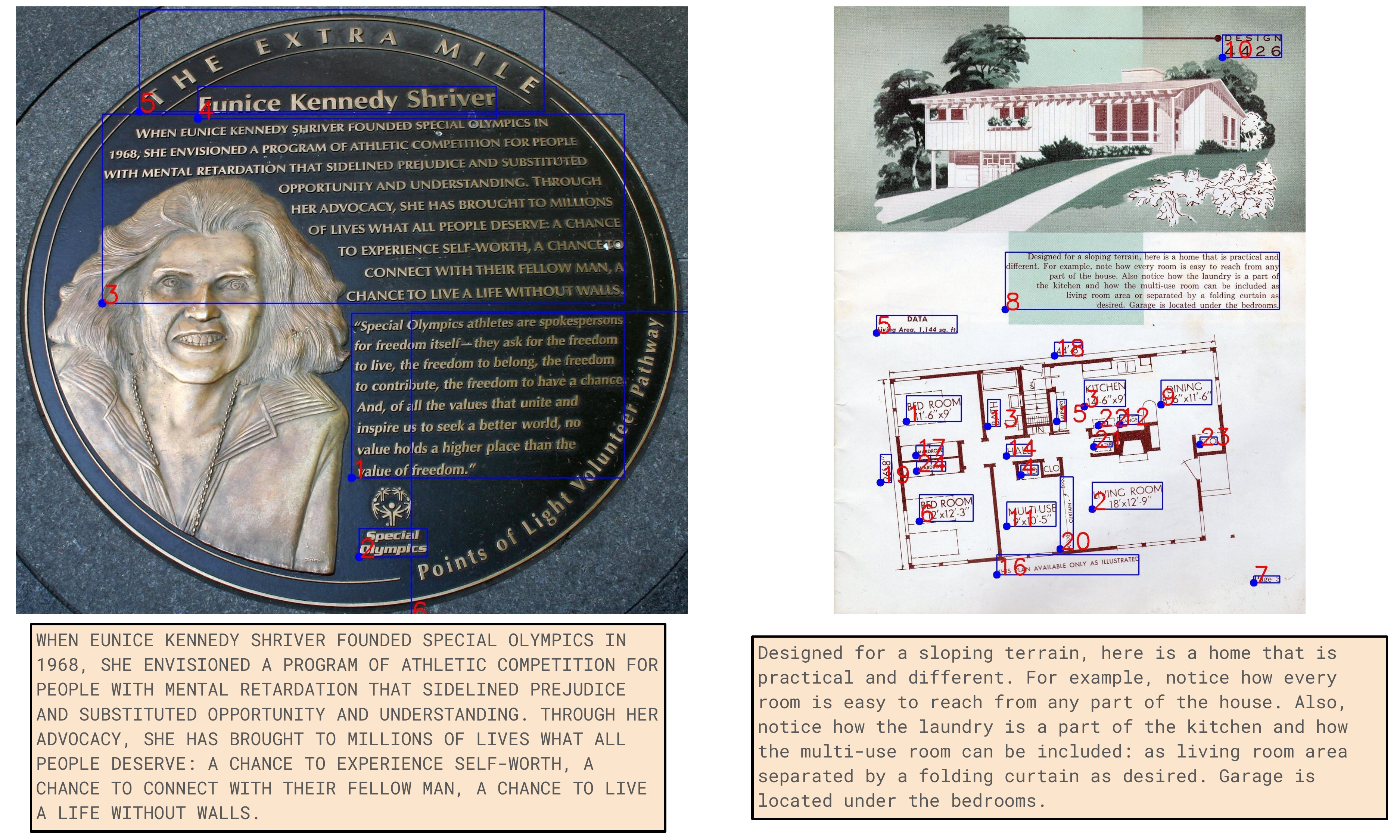}}
            \cprotect\caption{Examples outputs from \technique{} on images from the Hiertext dataset. For readability, only the text recognized for the biggest block in each image is given.}
            \label{fig:examples}
        \end{figure}

        The results\footnote{GPT-3.5 Turbo used when evaluating \technique{} was accessed on 2023-10-20} of \technique{} on our block-level text spotting dataset are given in Table \ref{table:results}. The Hiertext dataset is a challenging dataset for text spotting due to its high text density (see Section 3.2 in \cite{long2022endtoend}), yet \technique{} performs well on both learned metrics and string similarity metrics. Some qualitative results are shown in Figure \ref{fig:examples}.

        \begin{table}
            \centering
            \makebox[\textwidth][c]{
            \begin{tabular}{|c|c|c|}
            \hline
            \textit{Metric}                              & \textit{Result} \\ \hline
            Sentence Transformer $\uparrow$              & 0.7407          \\ \hline
            BERTScore $\uparrow$                         & 0.7862          \\ \hline
            BLEURT $\uparrow$                            & 0.6123          \\ \hline
            Jaro-Winkler Similarity $\uparrow$           & 0.8502          \\ \hline
            Ratcliff-Obershelp Similarity $\uparrow$     & 0.7870          \\ \hline
            Normalized Levenshtein Distance $\downarrow$ & 0.2614          \\ \hline
            \end{tabular}}
            \caption{Results of \technique{} on our block-level text spotting dataset. $\uparrow$ indicates that higher is better and $\downarrow$ indicates that lower is better.}
            \label{table:results}
        \end{table}

\section{Conclusion}
    We proposed a new technique, \technique{}, which incorporates an LLM into the pipeline for block-level text spotting. \technique{} was evaluated on our block-level text spotting dataset using both learned metrics (that measure semantic likeness) and string similarity metrics (that measure character-level likeness). It was seen to perform well on both types of metrics on our challenging block-level dataset.
    
    Incorporating an LLM into text spotting allows it to correct mistakes made during text recognition and reconstruct semantically meaningful text that was corrupted during recognition. The semantic and world knowledge contained in an LLM is vast, much larger than any text spotting model. Hence, incorporating it has the potential to improve the quality of text spotting which would consequently enhance performance on downstream tasks. For future work, the impact of the quality improvements provided by an LLM could be quantified by measuring the improvement in results on downstream tasks such as scene text VQA.

\printbibliography

\appendix

\section{Results of Retrained PARSeq Model} \label{appendix:rorpm}
    When retraining PARSeq, the data, training code and training parameters from \cite{bautista2022scene} were used without modification. Only two changes were made: 1) The space character was added to the vocabulary. 2) We skipped the preprocessing step of removing spaces in ground truth labels. Additionally, the autoregressive decoding scheme was used during inference. When comparing results of the retrained PARSeq to the original, see \subscript{PARSeq}{A} with \verb|Train data: R| in the original tables. Tables \ref{table:PARSeq_94}, \ref{table:PARSeq_62} and \ref{table:PARSeq_36} give the main results of the retrained PARSeq. Table \ref{table:PARSeq_challenging} gives results on longer and more challenging datasets. Tables \ref{table:PARSeq_90}, \ref{table:PARSeq_180} and \ref{table:PARSeq_270} give results on rotated datasets. %

    \begin{table}
        \centering
        \makebox[\textwidth][c]{
        \begin{tabular}{|c|c|c|c|c|c|}
        \hline
        \textit{Dataset}    & \textit{Number of Samples} & \textit{Accuracy} & \textit{1 - NED} & \textit{Confidence} & \textit{Label Length} \\ \hline
        IIIT5k              & 3000                       & 95.27             & 98.28            & 96.71               & 5.23                  \\ \hline
        SVT                 & 647                        & 95.36             & 98.59            & 95.65               & 5.88                  \\ \hline
        IC13\_1015          & 1015                       & 97.34             & 99.05            & 97.20               & 5.32                  \\ \hline
        IC15\_2077          & 2077                       & 85.27             & 94.57            & 91.42               & 5.43                  \\ \hline
        SVTP                & 645                        & 94.42             & 98.18            & 93.99               & 5.88                  \\ \hline
        CUTE80              & 288                        & 95.14             & 98.11            & 96.11               & 5.57                  \\ \hline \hline
        Combined            & 7672                       & \textbf{92.77}    & 97.39            & 95.00               & 5.42                  \\ \hline
        \end{tabular}}
        \caption{Accuracy of retrained PARSeq with the 95 character vocabulary (94 character original vocabulary + space character). Compare this to Table 4 in \cite{bautista2022scene}.}
        \label{table:PARSeq_94}
    \end{table}

    \begin{table}
        \centering
        \makebox[\textwidth][c]{
        \begin{tabular}{|c|c|c|c|c|c|}
        \hline
        \textit{Dataset} & \textit{Number of Samples} & \textit{Accuracy} & \textit{1 - NED} & \textit{Confidence} & \textit{Label Length} \\ \hline
        IIIT5k           & 3000                       & 96.47             & 98.48            & 96.71               & 5.09                  \\ \hline
        SVT              & 647                        & 96.75             & 98.86            & 95.65               & 5.86                  \\ \hline
        IC13\_1015       & 1015                       & 97.73             & 99.11            & 97.20               & 5.32                  \\ \hline
        IC15\_2077       & 2077                       & 87.10             & 94.89            & 91.42               & 5.33                  \\ \hline
        SVTP             & 645                        & 94.88             & 98.29            & 93.99               & 5.87                  \\ \hline
        CUTE80           & 288                        & 95.83             & 98.18            & 96.11               & 5.53                  \\ \hline \hline
        Combined         & 7672                       & \textbf{93.97}    & 97.59            & 95.00               & 5.33                  \\ \hline
        \end{tabular}}
        \caption{Accuracy of retrained PARSeq with the 63 character vocabulary (62 character original vocabulary + space character). Compare this to Table 4 in \cite{bautista2022scene}.}
        \label{table:PARSeq_62}
    \end{table}

    \begin{table}
        \centering
        \makebox[\textwidth][c]{
        \begin{tabular}{|c|c|c|c|c|c|}
        \hline
        \textit{Dataset} & \textit{Number of Samples} & \textit{Accuracy} & \textit{1 - NED} & \textit{Confidence} & \textit{Label Length} \\ \hline
        IIIT5k           & 3000                       & 97.70             & 99.28            & 96.71               & 5.09                  \\ \hline
        SVT              & 647                        & 97.37             & 99.08            & 95.65               & 5.86                  \\ \hline
        IC13\_1015       & 1015                       & 98.42             & 99.51            & 97.20               & 5.32                  \\ \hline
        IC15\_2077       & 2077                       & 89.22             & 96.28            & 91.42               & 5.33                  \\ \hline
        SVTP             & 645                        & 95.50             & 98.63            & 93.99               & 5.87                  \\ \hline
        CUTE80           & 288                        & 96.53             & 98.61            & 96.11               & 5.53                  \\ \hline \hline
        Combined         & 7672                       & \textbf{95.24}    & 98.40            & 95.00               & 5.33                  \\ \hline
        \end{tabular}}
        \caption{Accuracy of retrained PARSeq with the 37 character vocabulary (36 character original vocabulary + space character). Compare this to Table 6 in \cite{bautista2022scene}.}
        \label{table:PARSeq_36}
    \end{table}

    \begin{table}
        \centering
        \makebox[\textwidth][c]{
        \begin{tabular}{|c|c|c|c|c|c|}
        \hline
        \textit{Dataset} & \textit{Number of Samples} & \textit{Accuracy} & \textit{1 - NED} & \textit{Confidence} & \textit{Label Length} \\ \hline
        ArT              & 35149                      & 83.46             & 94.41            & 90.27               & 5.41                  \\ \hline
        COCOv1.4         & 9825                       & 79.28             & 92.47            & 83.59               & 5.90                  \\ \hline
        Uber             & 80382                      & 82.43             & 92.36            & 85.64               & 5.48                  \\ \hline \hline
        Combined         & 125356                     & \textbf{82.47}    & 92.95            & 86.77               & 5.49                  \\ \hline
        \end{tabular}}
        \caption{Accuracy of retrained PARSeq on larger and more challenging datasets using the 37 character vocabulary (36 character original vocabulary + space character). Compare this to Table 5 in \cite{bautista2022scene}.}
        \label{table:PARSeq_challenging}
    \end{table}

    \begin{table}
        \centering
        \makebox[\textwidth][c]{
        \begin{tabular}{|c|c|c|c|c|c|}
        \hline
        \textit{Dataset} & \textit{Number of Samples} & \textit{Accuracy} & \textit{1 - NED} & \textit{Confidence} & \textit{Label Length} \\ \hline
        IIIT5k           & 3000                       & 97.70             & 99.28            & 96.71               & 5.09                  \\ \hline
        SVT              & 647                        & 97.37             & 99.08            & 95.65               & 5.86                  \\ \hline
        IC13\_1015       & 1015                       & 98.42             & 99.51            & 97.20               & 5.32                  \\ \hline
        IC15\_2077       & 2077                       & 89.22             & 96.28            & 91.42               & 5.33                  \\ \hline
        SVTP             & 645                        & 95.50             & 98.63            & 93.99               & 5.87                  \\ \hline
        CUTE80           & 288                        & 96.53             & 98.61            & 96.11               & 5.53                  \\ \hline \hline
        Combined         & 7672                       & \textbf{95.24}    & 98.40            & 95.00               & 5.33                  \\ \hline
        \end{tabular}}
        \caption{Accuracy of retrained PARSeq on rotated versions of datasets using the 95 character vocabulary. Rotated by 90\textdegree. Compare this to Table 17 in Appendix J of \cite{bautista2022scene}.}
        \label{table:PARSeq_90}
    \end{table}

    \begin{table}
        \centering
        \makebox[\textwidth][c]{
        \begin{tabular}{|c|c|c|c|c|c|}
        \hline
        \textit{Dataset} & \textit{Number of Samples} & \textit{Accuracy} & \textit{1 - NED} & \textit{Confidence} & \textit{Label Length} \\ \hline
        IIIT5k           & 3000                       & 88.57             & 93.62            & 93.27               & 5.21                  \\ \hline
        SVT              & 647                        & 89.80             & 94.72            & 91.78               & 5.88                  \\ \hline
        IC13\_1015       & 1015                       & 92.02             & 94.61            & 94.36               & 5.31                  \\ \hline
        IC15\_2077       & 2077                       & 78.96             & 90.20            & 86.77               & 5.40                  \\ \hline
        SVTP             & 645                        & 83.72             & 90.46            & 87.04               & 5.84                  \\ \hline
        CUTE80           & 288                        & 87.50             & 92.28            & 93.28               & 5.51                  \\ \hline \hline
        Combined         & 7672                       & \textbf{86.08}    & 92.60            & 91.00               & 5.40                  \\ \hline
        \end{tabular}}
        \caption{Accuracy of retrained PARSeq on rotated versions of datasets using the 95 character vocabulary. Rotated by 180\textdegree. Compare this to Table 17 in Appendix J of \cite{bautista2022scene}.}
        \label{table:PARSeq_180}
    \end{table}

    \begin{table}
        \centering
        \makebox[\textwidth][c]{
        \begin{tabular}{|c|c|c|c|c|c|}
        \hline
        \textit{Dataset} & \textit{Number of Samples} & \textit{Accuracy} & \textit{1 - NED} & \textit{Confidence} & \textit{Label Length} \\ \hline
        IIIT5k           & 3000                       & 86.87             & 94.38            & 89.79               & 5.17                  \\ \hline
        SVT              & 647                        & 86.40             & 94.68            & 87.43               & 5.85                  \\ \hline
        IC13\_1015       & 1015                       & 90.25             & 95.05            & 91.76               & 5.31                  \\ \hline
        IC15\_2077       & 2077                       & 77.27             & 90.84            & 83.93               & 5.36                  \\ \hline
        SVTP             & 645                        & 79.84             & 92.28            & 84.33               & 5.83                  \\ \hline
        CUTE80           & 288                        & 82.99             & 91.33            & 87.60               & 5.43                  \\ \hline \hline
        Combined         & 7672                       & \textbf{83.94}    & 93.25            & 87.72               & 5.36                  \\ \hline
        \end{tabular}}
        \caption{Accuracy of retrained PARSeq on rotated versions of datasets using the 95 character vocabulary. Rotated by 270\textdegree. Compare this to Table 17 in Appendix J of \cite{bautista2022scene}.}
        \label{table:PARSeq_270}
    \end{table}

\section{Best Fuzzy Substring Match} \label{appendix:substring}
    The best fuzzy substring match of a string \verb|query| in a string \verb|corpus| is defined as the substring in \verb|corpus| which has the least Levenshtein distance from \verb|query|. The brute force method would be to calculate the Levenshtein distance between \verb|query| and every substring in \verb|corpus| and select the substring with the minimum distance. However this brute force approach is too slow and so our algorithm performs the search in two stages. A few preliminaries before describing the algorithm:\\
    \textit{Scan}: A scan moves a sliding window over the string being scanned and is parameterized by 1) window size and 2) step size (number of positions the window is moved in one step).\\
    \textit{Resolution of a scan}: A low resolution scan is one with a large step size and a high resolution scan is one with a small step size.
    
    Stage 1 is a low resolution scan over \verb|corpus| to find the region in which the best match may exist. Stage 1 will quickly narrow down the region which needs to be searched in \verb|corpus|. Stage 2 consists of multiple high resolution scans considering various window sizes over the narrowed down region. Stage 2 will perform a more thorough search in the narrowed down region to find the best match. The algorithm is not guaranteed to find the best match but it is empirically seen to find the best match in almost all normal strings\footnote{Abnormal strings are those with irregularities such as long sequences of whitespaces, a large number of repeating characters or character sequences, etc.}, while being significantly faster than a brute force approach.
    
\begin{figure}
        \centering
        \makebox[\textwidth][c]{\includegraphics[width=1.25\textwidth]{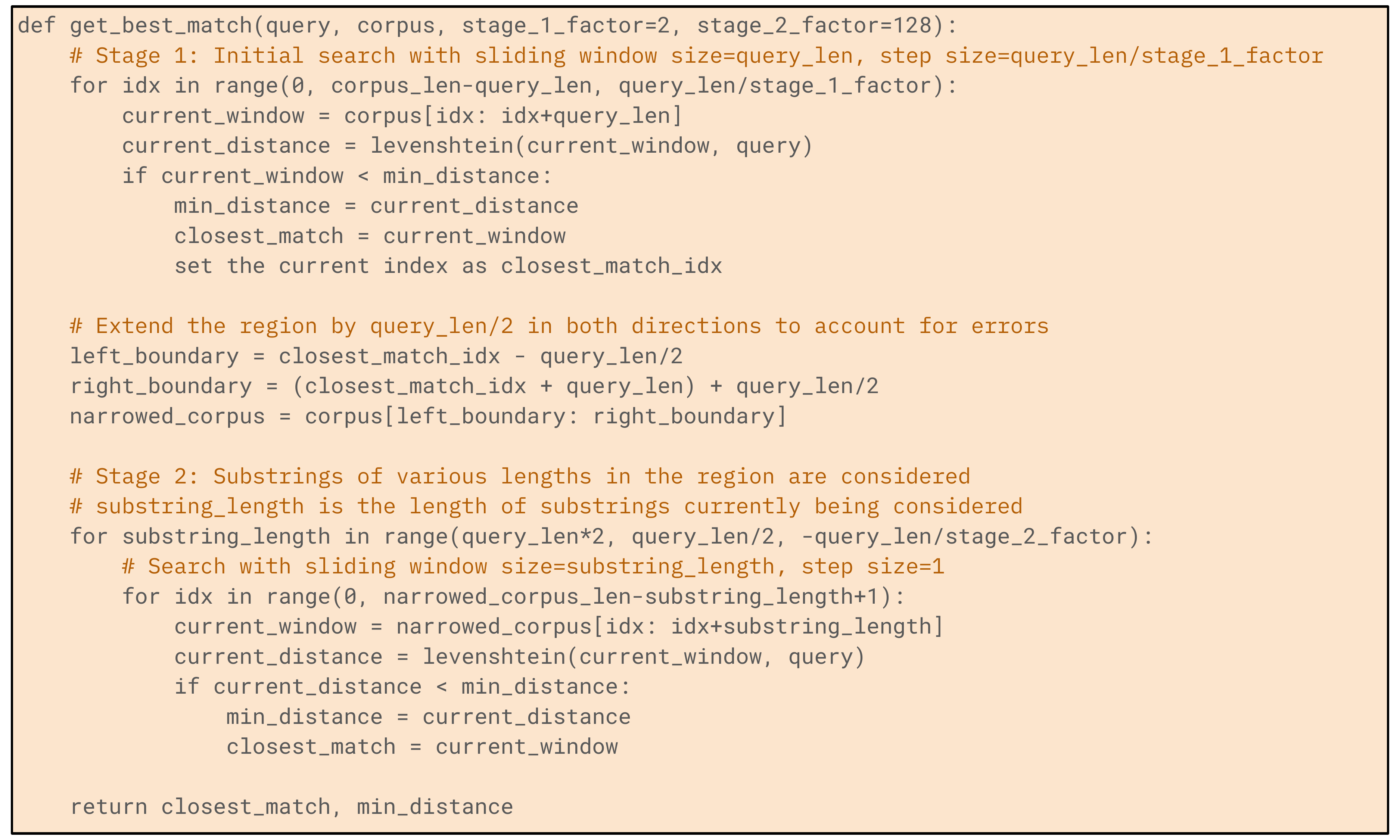}}
        \cprotect\caption{Algorithm for finding the best fuzzy substring match of \verb|query| in \verb|corpus|. Written in Python-style pseudocode.}
        \label{fig:substring_algorithm}
    \end{figure}

    The algorithm used to find the best fuzzy substring match is shown in Figure \ref{fig:substring_algorithm} as Python-style pseudocode. Apart from \verb|query| and \verb|corpus|, the arguments are:
    \begin{itemize}
        \item \verb|stage_1_factor|: It decides the step size of the scan in Stage 1
        \item \verb|stage_2_factor|: It decides the interval at which substring lengths will be considered in the multiple scans of Stage 2
    \end{itemize}
    
\end{document}